\documentclass[sigconf, nonacm]{acmart}
\makeatletter
\@ACM@balancefalse
\makeatother
\usepackage{pifont}
\usepackage{multirow}
\usepackage{booktabs}
\usepackage{tabularx}
\usepackage{balance}
\usepackage{afterpage}

\AtBeginDocument{%
  }

\begin{document}

\title{Wrong Code, Right Structure: Learning Netlist Representations from Imperfect LLM-Generated RTL}

\newcommand{\affone}{\texorpdfstring{\textsuperscript{1}}{}}
\newcommand{\affthree}{\texorpdfstring{\textsuperscript{3}}{}}
\newcommand{\affonethree}{\texorpdfstring{\textsuperscript{1,3}}{}}
\newcommand{\afftwothree}{\texorpdfstring{\textsuperscript{2,3}}{}}
\newcommand{\authorsep}{\texorpdfstring{\quad}{, }}
\author{Siyang Cai\affonethree\authorsep
Cangyuan Li\affthree\authorsep
Haoyu Gao\affonethree\authorsep
Kun Wang\afftwothree\authorsep
Yinhe Han\affthree\authorsep
Ying Wang\affthree}
\affiliation[obeypunctuation=true]{
  \institution{\affone School of Advanced Interdisciplinary Sciences, University of Chinese Academy of Sciences, Beijing, China\\
  \textsuperscript{2}Hangzhou Institute for Advanced Study, University of Chinese Academy of Sciences, Hangzhou, China\\
  \affthree CICS, Institute of Computing Technology, Chinese Academy of Sciences, Beijing, China\\
  \href{mailto:caisiyang23@mails.ucas.ac.cn}{caisiyang23@mails.ucas.ac.cn}\quad
  \href{mailto:wangying2009@ict.ac.cn}{wangying2009@ict.ac.cn}}
  \city{\relax}
  \country{\relax}
}

\renewcommand{\shortauthors}{Cai et al.}
\begin{abstract}

    Learning effective netlist representations is fundamentally constrained by the scarcity of labeled circuit data, as realistic designs are protected as intellectual property (IP) and costly to annotate. Existing task-specific studies therefore often focus on small-scale circuits with clean labels, limiting generalization to larger designs. Meanwhile, Large Language Models (LLMs) can generate Register-Transfer-Level (RTL) code at scale, but their functional incorrectness has hindered their use in circuit analysis. In this work, we make a key observation: even when LLM-Generated RTL is functionally imperfect, the synthesized netlists still preserve structural patterns that are strongly indicative of the intended functionality. Building on this insight, we propose a data augmentation framework that systematically transforms imperfect LLM-Generated RTL into training data for netlist representation learning, forming an end-to-end pipeline from automated code generation to downstream tasks. We evaluate the resulting training corpora using both task-specific and pretrained circuit representation models across operator-level and IP-level identification tasks. Results show that models trained on the resulting synthetic corpus generalize to unseen netlists and match or surpass models trained on scarce, functionally verified data.

\end{abstract}

\maketitle

\section{Introduction}

Netlist representation learning supports a range of downstream tasks, including IP piracy detection~\cite{GNN4IP_ICCAD_2021}, functional understanding~\cite{gen-eda, widegate, RELUT-GNN, fgnn2, fgnn}, reverse engineering~\cite{GNN-RE, DAGNN-RE, app-gnn}, and hardware security auditing~\cite{gnn4tj, trojan-guard, yu2021hw2vec, yasaei2022hardware}. It maps discrete circuit structures to continuous embeddings that capture latent design intent.

Current methods predominantly adopt supervised learning~\cite{GNN-RE, DAGNN-RE} or self-supervised contrastive learning~\cite{fgnn, fgnn2}. Both paradigms remain constrained by limited data availability. For functional identification tasks, acquiring large-scale netlists with annotated boundaries remains difficult due to IP protection, often confining existing evaluations to operator-level circuits. Consequently, existing models are often trained on small, structurally limited datasets, hindering generalization to larger and more diverse designs.
\begin{figure}[t]
    \centering
    \includegraphics[width=\linewidth]{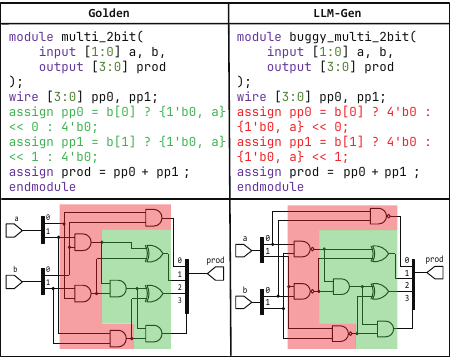}
    \caption{Functionally incorrect LLM-Generated RTL can retain structural features of the golden design after synthesis.}
    \Description{An example of functionally imperfect LLM-Generated RTL whose synthesized netlist retains structural similarity to the golden netlist.}
    \label{fig:intro}
\end{figure}

LLMs offer a promising path forward due to their demonstrated ability to generate RTL code at scale~\cite{rtlcoder, chang2024data, RTLLM_arXiv_2024}. However, LLM-Generated RTL is prone to functional errors, and establishing correctness requires costly verification and repair. As illustrated in Fig.~\ref{fig:intro}, a generated netlist can retain high structural similarity to the ground truth despite functional errors. This counterintuitive observation suggests that functional errors do not necessarily erase useful structural patterns, enabling noisy, unverified RTL code to serve as a scalable supervision signal for representation learning.

Translating this insight into a practical framework raises three challenges: (1) generating synthesizable RTL at scale; (2) distilling structurally reliable samples from a noisy, functionally imperfect corpus; and (3) maintaining sufficient diversity to generalize to unseen implementations.

In this work, we propose an end-to-end data augmentation and training framework that leverages LLM-Generated RTL for netlist representation learning. Without performing full functional verification for every candidate, we synthesize diverse, potentially imperfect RTL variants into gate-level netlists and convert them into graph representations. A structure-aware filtering mechanism leverages the structural and functional priors learned by a pretrained circuit encoder to retain structurally reliable samples from the noisy corpus. Furthermore, we encourage LLMs to generate architecturally distinct implementations, thereby improving the framework's generalization capability. We evaluate our framework across both operator-level and IP-level functional identification tasks.

We summarize our contributions as follows:

\begin{itemize}
    \item \textbf{Rethinking the Value of Imperfect RTL.} Through comprehensive studies, we show that functionally imperfect LLM-Generated RTL can retain useful structural patterns after synthesis, demonstrating its potential as training data for circuit representation learning.
    
    \item \textbf{Structure-Aware Data Augmentation Framework.} We introduce a structure-aware framework that repurposes functionally imperfect LLM-Generated RTL as a scalable source of supervision for netlist representation learning. Rather than treating functional correctness as a prerequisite for data utility, the framework distills structurally reliable and architecturally diverse training samples through synthesis-guided generation, structure-aware filtering, and architecture voting.
    
    \item \textbf{Scalability Across Circuit Scales and Backbones.} We extend subcircuit identification beyond operator-level arithmetic blocks to the IP level and demonstrate that the utility of the resulting training data persists across task-specific and pretrained circuit representation models.
\end{itemize}

\section{Related Work}
\subsection{Circuit Representation Learning}
Existing circuit representation learning approaches broadly follow two training paradigms: task-specific supervised learning and self-supervised pretraining. Task-specific approaches optimize circuit representations for downstream objectives, including subcircuit identification and reverse engineering~\cite{GNN-RE,DAGNN-RE,widegate}, as well as IP piracy and hardware Trojan detection~\cite{gnn4tj,GNN4IP_ICCAD_2021}. Pretraining approaches instead learn transferable structural and functional priors from simulation-derived signals, logic-equivalent views, or multimodal alignment~\cite{deepgate2,deepgate3,deepgate4,fgnn2,fang2025circuitfusion}. However, task-specific learning requires labeled circuits, whereas effective pretraining depends on large and structurally diverse corpora; both resources remain difficult to obtain at sufficient scale. In particular, subcircuit boundary identification requires gate-level labels indicating the functional module to which each gate belongs; recovering such labels manually from flattened netlists is costly, limiting the scale and coverage of supervised training data.
\subsection{Netlist Data Augmentation}
Conventional netlist augmentation explores the gate-level implementation space of a fixed RTL design through logic rewriting, synthesis optimization, and technology mapping~\cite{fgnn2,deepgate3,DAGNN-RE}. These transformations introduce useful structural variation while preserving functionality, but remain bounded by the architectural decisions already encoded in the source RTL.

Recent studies have advanced LLM-based RTL generation through specialized training corpora, automated data augmentation, benchmarks, and scalable generation pipelines~\cite{rtlcoder,chang2024data,RTLLM_arXiv_2024,Autosilicon}. Beyond scaling data volume, LLMs can generate architecturally distinct RTL implementations from functional specifications, enriching circuit corpora at a higher abstraction level than synthesis-based transformations. However, generated designs frequently contain functional errors, and how to select imperfect samples that still provide useful structural supervision for netlist representation learning has received limited attention.

\section{Methodology}
\begin{figure*}[t]
    \centering
    \includegraphics[width=\textwidth]{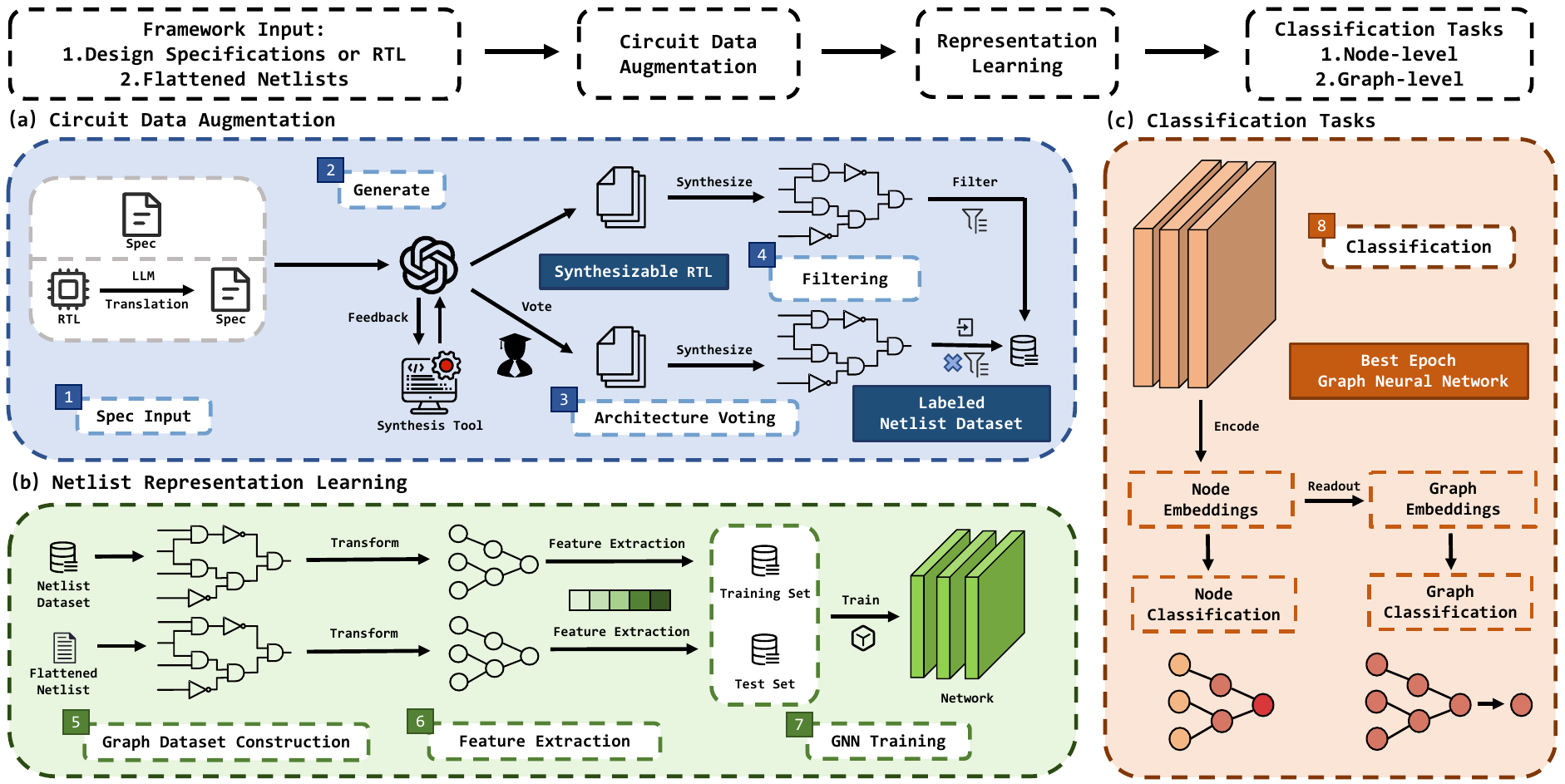}
    \caption{An overview of our proposed framework.}
    \Description{The framework generates RTL variants, synthesizes and filters their netlists, constructs graph datasets, learns netlist representations, and evaluates them on downstream classification tasks.}
    \label{fig:methodology}
\end{figure*}
\subsection{Framework Overview}
As shown in Fig.~\ref{fig:methodology}, our framework consists of three stages: \textbf{(a) Circuit Data Augmentation}, which generates RTL variants, synthesizes them into gate-level netlists, and curates the resulting candidates; \textbf{(b) Netlist Representation Learning}, which converts the resulting netlists into graph representations and learns circuit embeddings using a GNN backbone; and \textbf{(c) Classification Tasks}, which applies the learned representations to node- and graph-level functional identification.

\subsection{Circuit Data Augmentation}

As shown in Fig.~\ref{fig:methodology}(a), this stage expands limited design inputs into a curated gate-level netlist corpus through three components: (1) an LLM-based RTL generation pipeline that produces and synthesizes multiple candidate implementations; (2) a netlist-level filtering mechanism that retains candidates with sufficient structural similarity to the reference design; and (3) an RTL-level architecture voting mechanism that selects architecturally distinct candidates. The latter two components serve as complementary curation mechanisms for different generation objectives.

\subsubsection{LLM-based RTL Generation Pipeline}
The generation pipeline illustrated in Fig.~\ref{fig:methodology} accepts two forms of input: a user-provided functional specification or an existing RTL design. For the latter, the LLM first extracts a functional specification (\ding{172}), after which both input forms follow the same specification-driven generation process (\ding{173}). This unified workflow enables the LLM to generate architecturally diverse implementations of the same target functionality. By regenerating RTL from specifications rather than directly mutating existing code, we naturally introduce structural variations that maintain functional relevance without requiring hand-crafted mutation rules.

To support complex hierarchical designs, which remain challenging for LLM-based RTL generation~\cite{Autosilicon}, we adopt a bottom-up generation strategy. Each submodule is generated from its own functionally complete specification and then integrated at the top level, with rule-based checks applied to verify interface consistency and submodule instantiations. To improve RTL synthesizability, an additional debugging agent analyzes synthesis logs and guides RTL repair for up to three iterations. Candidates that still fail synthesis are discarded. For hierarchical designs, the same loop also checks for unresolved top-level references.

\subsubsection{Netlist-level Filtering Mechanism}
\label{sec:3.2.2}
Recognizing that augmented data can be unreliable, we implement a two-stage filtering mechanism at the netlist level to ensure the quality and relevance of our synthetic training data.

First, we remove candidates with unusually low post-synthesis cell counts. Candidates whose cell counts are more than one standard deviation below the mean of their functional class are discarded before structural screening.

Second, we apply structure-aware filtering based on the observation that LLM-Generated RTL, while potentially containing functional errors, can still produce netlists with structural patterns characteristic of the intended function. Structure-aware filtering is applied to candidates generated for reference-aligned augmentation, whereas candidates intentionally generated to explore alternative architectures are curated through the voting mechanism described in Sec.~\ref{sec:3.2.3}. To implement this screening, we compute a structural similarity score between a generated netlist ($G_{gen}$) and its corresponding golden netlist ($G_{gold}$). Both netlists are converted into graph representations and encoded using a frozen DeepGate4 encoder~\cite{deepgate4}, which leverages pretrained structural and functional priors to obtain node embeddings ($H_V$). We then construct a graph-level descriptor by concatenating the mean- and max-pooled node embeddings, i.e., $\mathbf{h}_G=\operatorname{MeanPool}(H_V)\,\Vert\,\operatorname{MaxPool}(H_V)$.

Accordingly, we compute the cosine similarity between the two graph-level descriptors:
\begin{equation}
\operatorname{Sim}(G_{gold},G_{gen})
=
\frac{\mathbf{h}_{G_{gold}}\cdot\mathbf{h}_{G_{gen}}}
{\lVert\mathbf{h}_{G_{gold}}\rVert\lVert\mathbf{h}_{G_{gen}}\rVert}.
\label{eq:filter_similarity}
\end{equation}

Here, $\mathbf{h}_G$ is a learned structural descriptor used to compare generated and reference netlists during candidate filtering. We define a similarity threshold $\tau$ and discard generated netlists with $\operatorname{Sim}(G_{gold},G_{gen})<\tau$. Rule-based augmentation methods, such as logic rewriting in FGNN2~\cite{fgnn2}, preserve functional equivalence through algebraic transformations, while their structural variations generally remain tied to the source implementation. Our filtering mechanism addresses a different generation setting by screening LLM-Generated candidates according to their learned structural similarity to the reference netlist. It retains candidates exhibiting structural cues associated with the target functionality while permitting variations in their global architectural organization. The threshold $\tau$ therefore controls the balance between structural relevance and architectural diversity.

\vspace{-0.45em}
\subsubsection{RTL-level Architecture Voting}
\label{sec:3.2.3}
LLM-based RTL generation can produce architecturally distinct implementations given the same functional specification, including alternative operator architectures, datapath organizations, and control structures. Incorporating these variants broadens the architectural coverage of the training corpus beyond transformations of a fixed source implementation.

Candidates intentionally generated to explore alternative architectures may receive low similarity scores under reference-based structure-aware filtering and thus be discarded. Motivated by prior work on LLM-based ranking and voting~\cite{voting, embodied, Autosilicon}, we therefore employ an LLM-guided architecture-voting strategy (Fig.~\ref{fig:methodology}, Step~\ding{174}) to prioritize candidates with greater architectural distinctiveness.

For each design task, the LLM ranks a batch of $N$ synthesizable candidates. Using the functional specification as a common reference, the ranking prompt prioritizes differences in operator architecture, datapath organization, and control structure. We select the top-$k$ candidates to enrich the training corpus with alternative implementations.

\subsection{Netlist Representation Learning}
Following data augmentation, the curated gate-level netlists enter the representation-learning stage shown in Fig.~\ref{fig:methodology}(b). We adopt GNN-RE~\cite{GNN-RE} as the primary representation-learning backbone of our framework. This stage consists of three components: graph dataset construction, feature extraction, and GNN training.

\subsubsection{Graph Dataset Construction}
At Step~\ding{176} in Fig.~\ref{fig:methodology}, we convert the augmented and available real-world flattened netlists into graph representations. Following GNN-RE, each netlist is represented as an undirected graph $G=(V,E)$, where $V$ denotes gates and $E$ denotes wires. Although logic circuits are directional, fan-in and fan-out information is retained through the input- and output-degree features of each node. Primary inputs (PIs) and primary outputs (POs) are encoded as attributes of their connected gates rather than as separate nodes.

Node-level functional classification requires each gate in a flattened netlist to be associated with its source functional module. We therefore employ a boundary-aware flattening flow. Module boundaries are preserved during the initial synthesis stage to establish gate-to-module correspondence, after which the design is flattened and the corresponding functional labels are propagated to the graph nodes. These labels provide the ground-truth annotations used for node-level training and evaluation.

\begin{figure}[t]
    \centering
    \includegraphics[width=\linewidth]{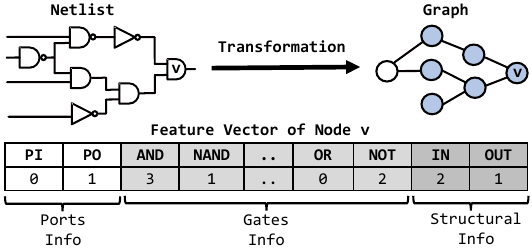}
    \caption{Netlist-to-graph transformation and feature extraction. Blue nodes mark gate $v$'s two-hop neighborhood.}
    \Description{A netlist is transformed into a graph with connectivity, local functional, and structural node features; blue nodes mark gate v's two-hop neighborhood.}
    \label{fig:node}
\end{figure}
\subsubsection{Feature Extraction}
\label{sec:3.3.2}
To initialize the GNN, each node $v\in V$ is assigned a feature vector $\mathbf{x}_v\in\mathbb{R}^d$ at Step~\ding{177} in Fig.~\ref{fig:methodology}. Following GNN-RE~\cite{GNN-RE}, $\mathbf{x}_v$ combines three types of information: (1) connectivity to primary inputs and outputs; (2) counts of different gate types within the two-hop neighborhood of $v$; and (3) structural features given by its fan-in and fan-out degrees. As illustrated in Fig.~\ref{fig:node}, these features capture both the local functional composition and direction-related structural information surrounding each gate. The complete feature matrix is denoted as $X_V\in\mathbb{R}^{|V|\times d}$.

\subsubsection{GNN Training}
Given the graph dataset and node features, we train the primary GNN-RE backbone using GraphSAINT~\cite{GraphSAINT_ICLR_2020} for scalable minibatch learning on large netlists. Specifically, the random-walk sampler (RWS) constructs each minibatch by performing random walks from a set of root nodes, allowing message passing to operate on sampled subgraphs rather than the full graph. For each sampled subgraph, neighborhood aggregation is restricted to the sampled neighbors $\widetilde{\mathcal{N}}(v)$. Starting from $\mathbf{h}_v^{(0)}=\mathbf{x}_v$, the GNN produces the final node representation $\mathbf{z}_v=\mathbf{h}_v^{(K)}$ after $K$ layers.

\subsection{Functional Classification Tasks}

As shown in Fig.~\ref{fig:methodology}(c), the learned high-dimensional representations are used for two downstream functional classification tasks: (1) node-level classification, which assigns a functional subcircuit label to each gate in a flattened netlist; and (2) graph-level classification, which predicts the functional category of an entire netlist graph.

\subsubsection{Node-Level Classification}
The goal of node-level classification is to perform subcircuit identification within a flattened netlist. Formally, given the set of final node embeddings $Z = \{\mathbf{z}_v \mid v \in V\}$ for a flattened graph $G_{flat}$, the task is to assign each node $v$ to its correct functional subcircuit class. This is achieved by passing each node's final embedding $\mathbf{z}_v$ through a shallow Multi-Layer Perceptron (MLP) followed by a $\text{softmax}$ activation function to obtain the classification probabilities.

\subsubsection{Graph-Level Classification}
The graph-level classification task is designed for component identification, where a collection of netlist graphs $\mathcal{D}_{target} = \{G_1, \dots, G_m\}$ must be assigned correct functional labels. To perform this task, a graph-level embedding $\mathbf{z}_G$ is first derived from the node embeddings $Z$ using a global $\text{READOUT}$ function:
\begin{equation}
\mathbf{z}_G = \text{READOUT}(\{\mathbf{z}_v \mid v \in V\}).
\end{equation}
This graph-level embedding $\mathbf{z}_G$ is then passed through a separate $\text{MLP}_{graph}$ classifier to predict the final functional label:
\begin{equation}
\hat{y}_G = \text{softmax}(\text{MLP}_{graph}(\mathbf{z}_G)).
\end{equation}

\section{Experiments}

\subsection{Experimental Setup}

The experiments follow the framework's logic: we first establish the utility of imperfect RTL, then assess architecture voting and structure-aware filtering, and finally test transfer across RTL architectures, representation backbones, and SoC designs.

Experiments use DeepSeek-V3.2-Exp and GLM-4.6 to generate RTL candidates with a sampling temperature of 0.8. All candidates follow the same synthesis and graph-construction flow based on \textit{Synopsys Design Compiler} and the TSMC 90\,nm standard-cell library. To obtain gate-level ground truth, synthesis is first performed with \textit{-no\_autoungroup}, preserving the correspondence between RTL modules and their synthesized cells. We then apply \textit{ungroup -all -flatten} to obtain flattened netlists for graph construction while propagating the module-derived ground-truth labels to their constituent gates.

We use GNN-RE as our primary representation-learning backbone and train it using the PyTorch implementation of GraphSAINT~\cite{GraphSAINT_ICLR_2020} on an NVIDIA A100 80\,GB GPU. We further evaluate a frozen DeepGate4 encoder with a shared lightweight classifier. We report F1-Micro for aggregate prediction quality and F1-Macro for class-balanced performance. To isolate the utility of the generated data from gains attributable to corpus scale, we control the training-data budget at the circuit level by matching the number of synthesized netlists across compared datasets unless otherwise specified.

\subsection{Operator-Level Evaluation}
\subsubsection{Utility of Imperfect LLM-Generated RTL}

We first ask whether a corpus constructed without exhaustive functional verification can still supervise netlist representation learning. We use the operator-level subcircuit boundary identification benchmark from GNN-RE, comprising 37 operator netlists and their RTL implementations, of which 22 designs form the training set. The corresponding RTL implementations use compact, direct arithmetic expressions such as \texttt{+}, \texttt{-}, and \texttt{*}. We preserve the original train/test split, operator-family composition, and bit-width partitioning. The original training designs are replaced by variants generated solely from training-side functional specifications.

We independently execute the complete augmentation pipeline five times to reduce the influence of generation randomness. Table~\ref{tab:data_aug_comparison1} reports aggregate statistics across the five runs as $LLM_{stat}$, together with the best observed batch, $LLM_{best}$. Across runs, the augmented corpora remain competitive with the verified-data baseline, indicating that they preserve useful supervision on average despite the absence of exhaustive functional validation. Under the matched circuit-level budget, the best batch exceeds the baseline by 1.13 and 3.64 points in F1-Micro and F1-Macro, respectively.

\begin{table}[htbp]
\centering
\caption{Performance comparison on operator-level subcircuit identification: baseline vs. LLM-Augmented datasets.}
\label{tab:data_aug_comparison1}
\resizebox{\columnwidth}{!}{
\begin{tabular}{c|c|c|c|c} 
\toprule
\textbf{Dataset} & \textbf{Circuits} & \textbf{Training Nodes} & \textbf{F1-Micro (\%)} & \textbf{F1-Macro (\%)} \\ 
\midrule
Baseline & 22 & 58,036 & 97.10 & 90.15 \\ 
\midrule
$LLM_{stat}$\textsuperscript{*} & 22 & 75,020--177,399 & 96.87 $\pm$ 1.57 & 89.92 $\pm$ 3.49 \\ 
$LLM_{best}$\textsuperscript{\dag} & 22 & 75,020 & \textbf{98.23} & \textbf{93.79} \\ 
\bottomrule
\multicolumn{5}{l}{\footnotesize \textsuperscript{*} Mean $\pm$ std.; nodes show the five-dataset range. \textsuperscript{\dag} Best observed batch.}
\end{tabular}
}
\end{table}

\subsubsection{Effectiveness of Architecture Voting}
To test transfer across RTL organizations, we retain the GNN-RE training protocol and replace its direct-operator test RTL with manually constructed, functionally equivalent implementations with diverse architectures, including array and Wallace-tree multipliers and ripple-carry and carry-lookahead adders.

We compare three training sets: \textit{Baseline}, which uses the original GNN-RE training data; \textit{LLM-Raw}, which uses synthesizable LLM-Generated samples without structure-aware filtering or architecture voting; and \textit{Voting}, which is constructed exclusively from the architecture-voting branch without applying structure-aware filtering. This design isolates the effect of architecture voting on cross-architecture generalization.

Table~\ref{tab:data_aug_comparison2} shows that raw generation already raises F1-Macro over the original corpus, consistent with broader exposure to alternative implementations. Architecture voting further improves F1-Micro and F1-Macro over LLM-Raw by 1.98 and 2.07 percentage points, respectively. Based on our audit, voting consistently ranks structurally reorganized variants ahead of direct operator-style implementations, indicating stable ranking behavior. The joint gain supports architecture voting as a mechanism for converting generated implementation diversity into supervision that transfers to unseen architectures.

\begin{table}[h]
\centering
\caption{Effect of architecture voting on cross-architecture operator-level generalization.}
\label{tab:data_aug_comparison2}

\resizebox{\columnwidth}{!}{
\begin{tabular}{c|c|c|c|c}
\toprule
\textbf{Dataset} & \textbf{Circuits} & \textbf{Training Nodes} & \textbf{F1-Micro (\%)} & \textbf{F1-Macro (\%)} \\
\midrule
Baseline & 22 & 58,036 & 92.78 & 89.63 \\
\midrule
LLM-Raw & 22 & 80,369 & 92.47 & 91.72 \\
Voting & 22 & 92,023 & \textbf{94.45} & \textbf{93.79} \\
\bottomrule
\end{tabular}
}
\end{table}

\subsubsection{Generalization Across Representation Backbones}

We further test whether the utility of the generated corpus transfers to a pretrained circuit representation model. We select DeepGate4~\cite{deepgate4}, a model pretrained at scale to learn transferable structural and functional circuit representations. We freeze its encoder to produce node embeddings and train only a task-adaptive classification head. \textbf{Original} uses the benchmark corpus, \textbf{LLM-Raw} expands its circuit count fivefold, and \textbf{LLM-Filtered} restores the original circuit budget through structural selection. All settings share the same test graphs, and Table~\ref{tab:deepgate4_backbone} reports mean and standard deviation over three classifier seeds.

\begin{table}[H]
\centering
\caption{Effect of structural curation under a frozen DeepGate4 encoder.}
\label{tab:deepgate4_backbone}
\setlength{\tabcolsep}{3.5pt}
\resizebox{\columnwidth}{!}{
\begin{tabular}{c|c|c|c|c}
\toprule
\textbf{Dataset} & \textbf{Circuits} & \textbf{Training Nodes} & \textbf{F1-Micro (\%)} & \textbf{F1-Macro (\%)} \\
\midrule
Original & 22 & 58,036 & $45.99\pm0.75$ & $35.31\pm0.91$ \\
LLM-Raw & 110 & 496,448 & $40.47\pm1.41$ & $39.32\pm0.48$ \\
LLM-Filtered & 22 & 64,679 & $\mathbf{48.49\pm0.21}$ & $\mathbf{42.82\pm0.14}$ \\
\bottomrule
\end{tabular}
}
\end{table}

Table~\ref{tab:deepgate4_backbone} shows that simply scaling the training data is insufficient: uncurated generations do not consistently improve both F1-Micro and F1-Macro, suggesting that additional samples may introduce structurally misaligned supervision. Structure-aware curation resolves this issue, enabling LLM-Filtered to achieve the strongest and most stable results using only about one-seventh as many training nodes as LLM-Raw. Its lower absolute performance relative to the task-specific GNN-RE model is consistent with the different adaptation settings: GNN-RE jointly optimizes its representations for boundary-identification labels, whereas the DeepGate4 encoder remains frozen and only a lightweight classifier is adapted. Under this constrained setting, the gains from LLM-Filtered indicate that structure-aware filtering selects supervision that better aligns with the pretrained representations.

\subsection{IP-Level Case Study}

\subsubsection{Experimental Setting}

To evaluate scalability and cross-design generalization, we study zero-shot transfer for IP-level subcircuit identification across two architecturally distinct open-source SoCs. PicoSoC~\cite{picosoc}, built around the PicoRV32 core~\cite{picorv32}, serves as the source SoC, whereas the NEORV32 SoC~\cite{neorv32} is held out for evaluation. The task identifies IP blocks spanning the CPU core and peripheral components such as the memory interface, SPI controller, and UART. We construct the training corpus exclusively from functional specifications extracted from PicoSoC and evaluate directly on the NEORV32 SoC without target-design supervision. Because these specifications capture interface semantics and intended behavior without exposing implementation details, this setting directly tests whether the resulting structural supervision transfers beyond the source architecture.

\subsubsection{Sensitivity Analysis of the Structural Similarity Threshold}

The threshold $\tau$ controls the structural selectivity of candidate filtering. Using a common PicoSoC-derived candidate pool, we evaluate sensitivity to $\tau$ in zero-shot IP-level subcircuit identification on the NEORV32 SoC. Threshold-free random selection from the same pool serves as an uncurated reference. We additionally include a \textbf{Rule-Based} baseline following FGNN2~\cite{fgnn2}, which generates variants through logic rewriting, synthesis constraints, and multi-target optimization. All settings select 20 candidates per source module across 12 modules, keeping the number of training graphs fixed. This experiment characterizes sensitivity to $\tau$ rather than selecting a target-tuned operating point.

\begin{table}[H]
\centering
\caption{Sensitivity of zero-shot IP-level transfer to the structure-aware filtering threshold.}
\label{tab:tau_sensitivity}
\setlength{\tabcolsep}{3.5pt}
\resizebox{\columnwidth}{!}{
\begin{tabular}{c|c|c|c|c}
\toprule
\textbf{Dataset} & \textbf{$\tau$} & \textbf{Training Nodes} & \textbf{F1-Micro (\%)} & \textbf{F1-Macro (\%)} \\
\midrule
Rule-Based & -- & 735,800 & 63.28 & 43.57 \\
\midrule
Random selection & -- & 639,575 & 56.90 & 46.38 \\
\midrule
\multirow{4}{*}{Similarity filtering}
 & 0.50 & 641,069 & 66.50 & 42.80 \\
 & 0.70 & 641,056 & 59.24 & 40.99 \\
 & 0.85 & 641,204 & 83.71 & 53.58 \\
 & 0.95 & 641,256 & \textbf{84.46} & \textbf{54.96} \\
\bottomrule
\end{tabular}
}
\end{table}

The sweep reveals two selectivity regimes. Loose thresholds produce inconsistent changes across F1-Micro and F1-Macro, indicating that admitting more generated variants does not necessarily improve supervision. Both metrics increase substantially at $\tau\geq0.85$, suggesting that filtering becomes effective after excluding candidates with substantial structural deviation. Random selection provides the uncurated reference, while the Rule-Based baseline reflects augmentation confined to a fixed source implementation. Together, these results show that the transfer benefit arises from sufficiently selective structural curation rather than from generated variation alone.

\subsubsection{Failure-Mode Analysis}

Downstream performance alone does not explain why some functionally incorrect candidates remain useful. We therefore analyze each error by its \emph{structural reach}, defined as the extent to which the semantic mismatch affects the synthesized dependency structure. Our analysis considers retained candidates from two representative modules: the data/memory-path module \textit{picorv32\_\allowbreak mem\_\allowbreak if} and the control-dominated module \textit{picorv32\_\allowbreak control\_\allowbreak core}. For candidates that pass the interface and synthesis checks, we compare their RTL with the corresponding golden implementations and trace the affected dataflow or control dependencies.

The two modules exhibit different structural reach. Memory-path errors remain local to the dominant address/data paths, whereas control errors propagate across states and outputs, altering state roles, fanout, and control--data dependencies. These cases identify structural reach as an empirical boundary of the \textit{Wrong Code, Right Structure} hypothesis. Local semantic errors can preserve class-discriminative organization; globally coupled control errors are more disruptive. The high-selectivity gains are consistent with filtering candidates whose relevant circuit organization remains intact.

\subsubsection{Data Contamination Analysis}
A potential concern is that the observed cross-design transfer performance may reflect target-design leakage caused by the LLM reproducing open-source implementations memorized during pretraining. We therefore audit the generated variants against both the PicoSoC source and the unseen NEORV32 SoC target, using MinHash similarity over normalized RTL to measure textual overlap and cosine similarity between synthesized-netlist descriptors to assess language-independent structural correspondence.

\begin{table}[h]
\caption{Similarity analysis across LLM-Generated variants.}
\label{tab:leakage_results}
\centering
\resizebox{1\columnwidth}{!}{
\begin{tabular}{lccc}
\toprule
\textbf{Metric (Mean)} & \textbf{vs. Source\textsuperscript{*}} & \textbf{vs. Target\textsuperscript{\dag}} & \textbf{Source--Target ($\Delta$)} \\
\midrule
Text-level (MinHash)   & 0.1375 & 0.0002 & +0.1373 \\
Netlist-level (Cosine) & 0.8446 & 0.7720 & +0.0726 \\
\bottomrule
\multicolumn{4}{l}{\footnotesize \textsuperscript{*} PicoSoC
(Verilog), \textsuperscript{\dag} NEORV32 SoC (VHDL)} \\
\end{tabular}
}
\end{table}

Table~\ref{tab:leakage_results} shows that the generated variants exhibit negligible text overlap with the NEORV32 SoC and remain closer to PicoSoC under both measures. This source-proximal pattern argues against direct reproduction of the unseen target while indicating that the variants retain transferable circuit structure.

\section{Conclusion}
This work transforms functionally imperfect LLM-Generated RTL into curated supervision for circuit representation learning. By combining synthesis-guided generation, structure-aware filtering, and architecture voting, the framework retains task-relevant structure and implementation diversity without exhaustive functional validation. Experiments on operator-level and IP-level subcircuit identification show that the curated corpora remain competitive with verified data and deliver consistent gains across task-specific and pretrained backbones. These results demonstrate that effective curation, not data scale alone, makes imperfect RTL a practical source of supervision for circuit representation learning.

\afterpage{\balance}
\bibliographystyle{ACM-Reference-Format}
\bibliography{Mybib}

\end{document}